\documentclass[sigconf]{acmart}

\AtBeginDocument{%
  }

\copyrightyear{2023} 
\acmYear{2023} 
\setcopyright{acmlicensed}\acmConference[WWW '23 Companion]{Companion Proceedings of the ACM Web Conference 2023}{April 30-May 4, 2023}{Austin, TX, USA}
\acmBooktitle{Companion Proceedings of the ACM Web Conference 2023 (WWW '23 Companion), April 30-May 4, 2023, Austin, TX, USA}
\acmPrice{15.00}
\acmDOI{10.1145/3543873.3587687}
\acmISBN{978-1-4503-9419-2/23/04}

\usepackage[utf8]{inputenc} 
\usepackage[T1]{fontenc}    
\usepackage{hyperref}       
\usepackage{url}            
\usepackage{booktabs}       
\usepackage{amsfonts}       
\usepackage{nicefrac}       
\usepackage{microtype}      
\usepackage{xcolor}         

\usepackage{graphicx}
\graphicspath{ {./media/} }
\usepackage{float}
\usepackage{caption}
\usepackage{subcaption}

\begin{document}

\title[A Federated Learning Benchmark for Drug-Target Interaction]{A Federated Learning Benchmark for Drug-Target Interaction}

\author{Gianluca Mittone}
\authornotemark[1]
\email{gianluca.mittone@unito.it}
\orcid{0000-0002-1887-6911}
\affiliation{
  \institution{University of Turin}
  \city{Turin}
  \country{Italy}
}
\author{Filip Svoboda}
\authornote{Both authors contributed equally to this research.}
\email{fs437@cam.ac.uk}
\orcid{0000-0002-3566-2341}
\affiliation{
  \institution{University of Cambridge}
  \city{Cambridge}
  \country{UK}
}
\author{Marco Aldinucci}
\email{marco.aldinucci@unito.it}
\orcid{0000-0001-8788-0829}
\affiliation{
  \institution{University of Turin}
  \city{Turin}
  \country{Italy}
}
\author{Nicholas D. Lane}
\email{ndl32@cam.ac.uk}
\orcid{0000-0002-2728-8273}
\affiliation{
  \institution{University of Cambridge}
  \city{Cambridge}
  \country{UK}
}
\author{Pietro Lio'}
\email{pl219@cam.ac.uk}
\orcid{0000-0002-0540-5053}
\affiliation{
  \institution{University of Cambridge}
  \city{Cambridge}
  \country{UK}
}

\renewcommand{\shortauthors}{Mittone, Svoboda, et al.}

\begin{abstract}
    Aggregating pharmaceutical data in the drug-target interaction (DTI) domain can potentially deliver life-saving breakthroughs. It is, however, notoriously difficult due to regulatory constraints and commercial interests \cite{hie2018realizing, topol2019high}. This work proposes the application of federated learning, which is reconcilable with the industry's constraints. It does not require sharing any information that would reveal the entities' data or any other high-level summary. When used on a representative GraphDTA model and the KIBA dataset, it achieves up to 15\% improved performance relative to the best available non-privacy preserving alternative. Our extensive battery of experiments shows that, unlike in other domains, the non-IID data distribution in the DTI datasets does not deteriorate FL performance. Additionally, we identify a material trade-off between the benefits of adding new data and the cost of adding more clients.
\end{abstract}

\keywords{Federated Learning, Graph Neural Networks, Drug-Target Interaction, Benchmark}

\begin{center}
The following paper is the accepted version of ACM copyrighted material
\\[12pt]
\textit{Mittone, G., Svoboda, F., Aldinucci, M., Lane, N., \& Lió, P. (2023, April). A Federated Learning Benchmark for Drug-Target Interaction. In Companion Proceedings of the ACM Web Conference 2023 (pp. 1177-1181)}
\\[12pt]
presented at the WWW'23 conference in Austin, Texas.
\\[12pt]
DOI: \href{https://doi.org/10.1145/3543873.3587687}{10.1145/3543873.3587687}
\end{center}

\maketitle

\section{Introduction}

Federated learning (FL) is a privacy-preserving distributed learning that has gathered ground in healthcare applications over the past few years. Since it fits very well with the requirement of preserving patient data confidentiality, it saw considerable uptake in the analysis of Electronic Health Records and healthcare IoT, such as mobile health \cite{pfitzner2021federated, nguyen2022federated, joshi2022federated}. The closest application of Federated Learning to Drug-Target Interaction (DTI) was the solitary example of FL-QSAR, which presented the first federated model trained for a related drug discovery task, but stopped short of analyzing its performance beyond demonstrating its feasibility for up to 4 clients \cite{chen2021fl}. Instead, providing privacy and security to drug discovery in general, and DTI in particular, has been approached as a cryptography problem by obscuring the underlying data such that data itself and high-level statistics were rendered useless. However, a model was still trainable on it \cite{ma2020secure}. 

This paper delivers the first-ever Federated Learning benchmark for the DTI task, achieving up to a 15.53\% reduction in MSE compared to a possible ensemble learning-based alternative. Furthermore, we develop a novel comprehensive analysis framework for FL applications letting us identify and explain a significant and material difference between the sensitivity of FL to non-IID data in the DTI task and sensitivity to it in any other task FL has been previously applied to, and discover and explore the importance of data ownership structure in FL for DTI as a major performance determinant and a key consideration when engaging real-world actors in the process of cooperatively training models. Ours is a novel and comprehensive analysis of FL in a critical and under-explored data domain.

This paper's scope is limited to the drug-target interaction DTI task of the drug discovery domain due to computational and other practical considerations. This task regresses the tuple protein-drug input onto a vector describing their interaction. In this domain, we chose to work with a single representative model. We chose the GraphDTA \cite{nguyen2021graphdta} model as it is the backbone of many current state-of-the-art models \cite{huang2020deeppurpose, nguyen2021graph, elinas2020variational}. Our experiments aim to represent the complexities a federation of pharmaceutical labs would entail as realistically as possible. In particular, we deliberately explored the whole spectra of IID-ness and data ownership distribution. Finally, we only perform our experiments using the core algorithms in FL and distributed learning. This choice does not imply loss of generality, as any specific feature that might improve the performance of either one of them can be straightforwardly re-implemented for use by the other.

In summary, our contributions are the following:
\begin{itemize}
    \item we deliver the first-ever Federated Learning benchmark for the DTI task;
    \item we achieve up to a 15.53\% reduction in MSE when compared to a bagging-based alternative;
    \item we develop a novel comprehensive analysis framework for FL applications, allowing us to identify data ownership as a major performance determinant;
    \item we report almost 200 FL training results through many heatmaps characterising the performance of the final model when trained under a wide spectrum of non-IID-ness levels in data distribution and different federation sizes.
\end{itemize}
Each reported experiment needs approximately one GPU hour on a NVIDIA A40 or 7-8 on GTX-1080. We ran 197 experiments for 1,576 GPU hours spent on this research work.
\section{Methodology of the proposed approach}
\label{sec:approach}



This work proposes to use Federated Learning based on the FedAverage \cite{fedAvg} algorithm to fit the GraphDTA model \cite{nguyen2021graphdta} on the KIBA \cite{kiba} dataset split among multiple clients. 

\textbf{Federated learning} is a distributed learning paradigm that shares model parameters at a much lower frequency than standard distributed learning. The exchanged model weights conceal each client's data sufficiently to preclude reconstruction. Straightforward extensions permit further increases in data protection, and defences against other potential interference \cite{li2020federated, kairouz2021advances}.

\textbf{The KiBA dataset} reports 246,088 Kinase Inhibitor BioActivity (KIBA) scores for 52,498 chemical compounds and 467 kinase targets, originating from three separate large-scale biochemical assays of kinase inhibitors. The score is a superior aggregate metric derived from a previously utilised battery of measurements such as  IC$_{50}$, K$_{i}$, and K$_{d}$ \cite{kiba}. 


\textbf{The GraphDTA model} regresses the drug-target pair onto a continuous measurement of binding affinity for that pair, the KIBA score. It encodes the target as a 1D sequence and the drug as a molecular graph, making it possible for the model to capture the bonds among atoms directly \cite{nguyen2021graphdta}. To stay true to the simple the better ethos of this paper, we refrained from implementing fancy aggregation strategies or specialising too much the chosen model to fit the federated task. We limited ourselves to replacing stateful objects (outside the weights, clearly) with stateless ones: batch normalisation with layer normalisation and ADAM with SGD. These choices let us obtain more stable learning curves and cleaner convergence of the federated model without harming its performance. Both FL and non-FL ran with the same adjusted architecture.

\textbf{Our implementation} uses the open-source FLOWER\cite{beutel2020flower} framework to implement the model federation and to simulate its running on multiple clients. We use the FedAverage aggregation algorithm, which combines local stochastic gradient descent (SGD) on each client with a server that performs model averaging \cite{fedAvg}.

\textbf{The experimental setup} builds on the experiments usually associated with Federated Learning benchmarks while substantially expanding them. First, the model is compared against a suitable alternative. Given the lack of prior work, there was no ready candidate for this comparison. A centralised model is unsuitable since its use is unrealistic due to the aforementioned regulatory and commercial considerations. The cryptographic approaches to data anonymisation would be usable in real life; however,  they are not a direct competitor to Federated Learning. They can augment each other and provide joint solutions similar to what Federated learning with differential privacy does \cite{wei2020federated}. Ultimately, as a possible fair comparison, we chose a simple Bergman's ensemble \cite{breiman1996bagging} of models, each being trained separately on a different data split of the entire dataset. The data splits are maintained constants in comparing FL and Ensemble Learning. The choice of baseline algorithms for both FL and the ensemble is deliberate, as any extension applicable to one can be straightforwardly re-engineered for use with the other \cite{polato2022boosting}. Therefore, working with simple implementations provides us with a fair, uncoloured comparison of the two approaches rather than of their two randomly chosen extensions. The metric used to evaluate each experiment is the Mean Squared Error (MSE); in the case of FL, the MSE of the global model is computed, while in the case of bagging, the MSE of the ensemble is taken into account. The test set is the same for all experiments, allowing for a fair comparison of different runs.


\textbf{The code is made available on GitHub }\footnote{\url{https://github.com/Giemp95/FedDTI}}. It can be used out of the box without the knowledge of distributed or Federated learning. It works with PyTorch deep models, but it will eventually be compatible with TensorFlow. It is being shared for the benefit of the Biologists working on DTI and those interested in proving and capitalising on Federated Learning's usefulness as a secure, privacy-preserving, and performance-conserving platform for sharing pharmaceutical data under regulatory and commercial constraints.

\section{Results}

\textbf{Superior and privacy-preserving} performance of our network is displayed in table \ref{table:model_results}. It reports the performance difference between the federation of deep model architectures and an ensemble of the same architecture. Based on our experimental setup (Section \ref{sec:approach}), all experiments in this section exploit the same GraphDTA \cite{nguyen2021graphdta} architecture, and we consider our model's performance to be successful if it can match that of the non-private distributed alternative.

 
    

\begin{table*}[htp]
    \centering
    \caption{Performance of our DTI-FL relative to ensemble alternative \cite{breiman1996bagging}. The \% difference columns refer to the federated values compared to the ensemble ones; note that a positive difference in the MSE highlights worse learning performance, while negative differences indicate better MSE and, consequently, better learning performance.}
    \label{table:model_results}
    \begin{tabular}{r@{\hspace{12pt}}
    r@{\hspace{12pt}}
    r@{\hspace{12pt}}
    l@{\hspace{12pt}}
    c
    r@{\hspace{12pt}}
    r@{\hspace{12pt}}
    l@{\hspace{12pt}}}
    \hline
    \multicolumn{1}{c}{} & \multicolumn{3}{c}{\bf IID distribution} && \multicolumn{3}{c}{\bf non-IID distribution} \\
    \cmidrule{2-4} \cmidrule{6-8}
    \textbf{Client count} & \textbf{Ensemble MSE} & \textbf{Federated MSE} & \textbf{\% difference} && \textbf{Ensemble MSE} & \textbf{Federated MSE} & \textbf{\% difference} \\
    \midrule
     2 clients & 0.509 & 0.530 & +4.08\% && 0.550 & 0.556 & +1.19\% \\ 
     4 clients & 0.563 & 0.577 & +2.58\% && 0.556 & 0.556 & -0.05\% \\
     8 clients & 0.567 & 0.574 & +1.30\% && 0.568 & 0.574 & +1.20\% \\
     16 clients & 0.576 & 0.578 & +0.42\% && 0.573 & 0.578 & +0.690\% \\
     32 clients & 0.709 & 0.599 & -15.53\% && 0.579 & 0.578 & -0.024\% \\
     \hline
    \end{tabular}
\end{table*}

The results in table \ref{table:model_results} show that our approach can retain up to 15\% better performance relative to the distributed alternative while ensuring that no data or any other high-level summary of it is revealed \cite{beutel2020flower}. The general trend in the IID results points to a relative advantage for the ensembles at very low client counts that quickly dissipates, turns into parity, and from 16 clients up fully reverses as the client count increases. Second, the non-IID data display effective parity practically at all client counts, indicating that FL can deal with unequal data distributions much better than the distributed alternative. This matched performance, alongside FL's solid privacy and security guarantees \cite{beutel2020flower} entirely lacking in the distributed alternative, makes it a clear favourite for future distributed learning research in the DTI domain. Furthermore, the results invite us to explore deeper. In particular, seeing that the IID and non-IID performances are effectively matched, we ask how the FL performance develops under varying non-IID conditions in the following subsection.

\textbf{DTI is a data distribution-agnostic domain.} Data non-IID-ness in DTI is two-dimensional as there are two model inputs. The protein and the chemical are jointly taken in to predict their interaction. Consequently, we can investigate the distribution one dimension at a time, either non-IID to the protein or chemical inputs, or explore it in both dimensions simultaneously. Neither of these three approaches can be ruled out as a priory as the input classes are statistically independent of each other. Consequently, the domain does not lend itself easily to the established notions of non-IID-ness in FL, and we have to test non-IID-ness under all three conditions. 

Our experiments investigate the entire continuum of IID-ness rather than just its two extrema. IID data distribution is a random draw; each data point has an equal chance of being owned by each client. A non-IID distribution, on the other hand, assigns either proteins or drugs to specific clients, and these clients then own all experiments that contain said protein or drug. In the real world, these would be the laboratories looking for drugs targeting a specific protein or investigating the effects of a specific drug. 

\begin{figure}[ht]
  \centering
  \begin{subfigure}[t]{.42\textwidth}
    \centering
    \includegraphics[width=\textwidth]{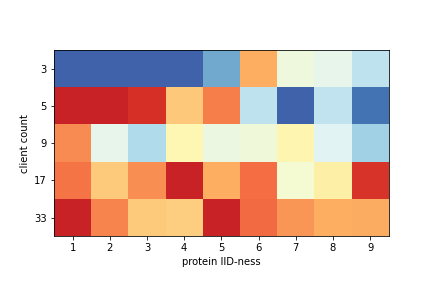}
    \Description{Protein-based IID-ness variation.}
    \caption{Protein-based IID-ness variation.}
    \label{fig:protein}
  \end{subfigure}
  \begin{subfigure}[t]{.42\textwidth}
    \centering
    \includegraphics[width=\textwidth]{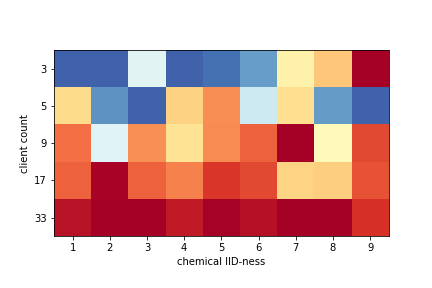}
    \Description{Chemical-based IID-ness variation.}
    \caption{Chemical-based IID-ness variation.}
    \label{fig:drug}
  \end{subfigure}
  \begin{subfigure}[t]{.42\textwidth}
    \centering
    \includegraphics[width=\textwidth]{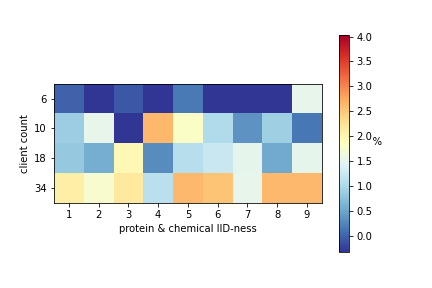}
    \Description{Combined IID-ness variation.}
    \caption{Combined IID-ness variation.}
    \label{fig:both}
  \end{subfigure}
  \Description{A \% change in MSE relative to the smallest client count and highest concentration in each setup is reported for a broad spectrum of client counts against (a) protein-based, (b) chemical-based, (c) protein- and chemical-based IID-ness variation. The horizontal axis represents the two extrema and seven equidistant points between them, vertical represents client count.}
  \caption{A \% change in MSE relative to the smallest client count and highest concentration in each setup is reported for a broad spectrum of client counts against (a) protein-based, (b) chemical-based, (c) protein- and chemical-based IID-ness variation. The horizontal axis represents the two extrema and seven equidistant points between them, vertical represents client count.}
  \label{fig:heat_maps}
\end{figure}

We obtained each row of each map in Figure \ref{fig:heat_maps} by first assigning to each client all experiments corresponding to an exclusive collection of either proteins or drugs. Then, at each step along the continuum, we let the clients exchange some of their data with their neighbours. This exchange follows a Gaussian curve, so we introduce an uneven representation of each data class outside its assigned client. This choice makes the distribution more realistic since it is unlikely that all clients but one would hold the same amount of data in any given class. We achieve the desired mix of protein- and drug-centric clients for the protein and drug experiments by splitting the data into two sub-datasets and then treating each as a separate one-class non-IID experiment. This scenario is closest to what we can expect in the real world. Each square in the figure reports the average over ten training iterations of the given model's loss performance relative to the centralized case. The client counts presented in these figures reflect the cross-silo setup of this domain.

Figures \ref{fig:protein}, \ref{fig:drug}, and \ref{fig:both} show the heat maps exploring the IID-ness space along the protein, drug, and both dimensions, respectively. As expected, having a higher client count hurts the performance at all non-IID-ness levels. That is, the more fragmented the dataset is, the more challenging the task of aggregating it, as the larger client count implies fewer data per client in this setup, which hurts the individual client models. The different levels of IID-ness, however, do not appear to have a link to the model's performance. In other words, while we see a general trend towards worse performance in each column, we do not see any such trend in the rows. This property is exciting, as it implies that it does not matter whether all client labs test the same combination of proteins or if each client has their own or substantially similar portfolio. It also means that what is a significant drain on FL's robustness in other domains is not a factor in the DTI domain.

In summary, unlike in other domains in which Federated Learning has been investigated, in the Drug-Target interaction, due to its unique data structure, the input IID-ness does not play a significant role, making the domain singularly unique among FL domains. This observation is crucial as resilience to non-IID data distribution is usually the chief robustness metric for comparing different aggregation strategies in FL. With the data distribution eliminated as a major limitation to our implementation's robustness, we turn to data quantity distribution, i.e. uneven data ownership, as the next candidate for a significant performance driver.

\begin{figure}[ht]
  \begin{subfigure}[t]{.42\textwidth}
    \centering
    \includegraphics[width=\textwidth]{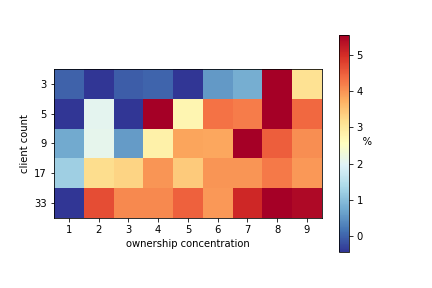}
    \Description{Data quantity IID-ness variation.}
    \caption{Data quantity IID-ness variation.}
    \label{fig:ownership}
  \end{subfigure}
  \hfill
  \begin{subfigure}[t]{.42\textwidth}
    \centering
    \includegraphics[width=\textwidth]{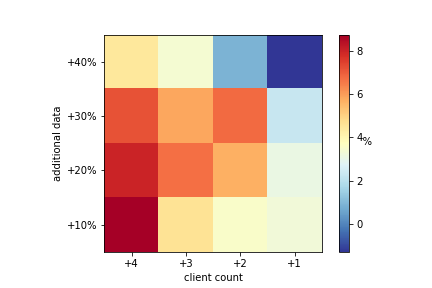}
    \Description{Data quantity IID-ness variation over different numbers of clients.}
    \caption{Data quantity IID-ness variation over different numbers of clients.}
    \label{fig:weakScaling}
  \end{subfigure}
  \Description{a): A \% change in MSE relative to the smallest client count and the highest concentration is reported for a selection of client counts and a range of data quantity distributions sampled equidistantly. b): A \% change in MSE relative to training solely based on the dominant client's (60\% of the) data is reported for the combinations of adding up to 40\% of extra data in increments of 10\% and divided among 1 to 4 additional clients.}
  \caption{a): A \% change in MSE relative to the smallest client count and the highest concentration is reported for a selection of client counts and a range of data quantity distributions sampled equidistantly. b): A \% change in MSE relative to training solely based on the dominant client's (60\% of the) data is reported for the combinations of adding up to 40\% of extra data in increments of 10\% and divided among 1 to 4 additional clients.}
  \label{fig:quantity}
\end{figure}

\textbf{Data distribution imbalance} plays a major role in Federated Learning's performance at DTI. Data distribution imbalance and unevenness in the data quantity among clients are of particular concern in the DTI domain, as the participant landscape is composed of a hodgepodge of big and small entities. The often-made assumption that clients have access to about the same amount of data, while plausible in some domains, is contrary to the structure of the pharmaceutical industry. Moreover, when this assumption is relaxed, it is argued that exploiting client size will speed up the training process, while data quantity distribution among the clients will ultimately not impact the model performance \cite{wu2021fast}. Figure \ref{fig:quantity} challenges this assumption and examines data quantity distribution's impact on the model performance under varying client counts.

Figure \ref{fig:ownership} investigates the interplay between client count and data quantity distribution profile. The dataset is distributed among multiple clients. The same single client is designated as the dominant client and receives a variable percentage of the data. The rest of the data is distributed unevenly among the rest of the clients following the Gaussian curve. This is done to achieve a reasonable uneven distribution in line with our approach exposed in the previous subsection. 

As before, increasing the client count makes the problem harder, increasing the error. This time, however, the rate of performance deterioration depends on the unevenness of data allocation among the clients. At each client count, irrespective of the ownership inequality level, it holds that moving to a more concentrated data ownership favours the model's performance. This effect is significant throughout the tested conditions but grows stronger the closer the tested setup is to the highly centralized data ownership.

Crucially, the co-dependent effect is not only present in the overwhelmingly dominant client case (far left), where it could be discounted as a case of mode collapse into a pseudo-centralized setup, but it holds throughout the tested conditions. This persistence makes our observation particularly salient. There is a cost to having a diluted client data ownership structure. Our next step is to investigate the interplay of this cost with the benefit of adding new data.

Figure \ref{fig:weakScaling} investigates the trade-off between the benefit of adding more data to an existing federation and the cost resulting from increasing the client count and thus diluting the client data ownership structure. We start with a single client allocated a 60\% share of the data. Without the loss of generality, this can represent a preexisting federation of clients. The remaining 40\% of the dataset is available for addition. The heat map reports the error implications from adding this data in increments of 10\% distributed among 1 to 4 clients. 

Predictably, increasing the amount of additional data and spreading this data among fewer clients improve model performance in Figure \ref{fig:weakScaling}. What is less predictable is that the rate of improvement is about the same in both of these dimensions, which is indeed remarkable. In the tested situation, increasing the concentration of data ownership can, in some cases, have as strong a positive effect on the model's performance as adding 10\% of the data. Consequently, we see that the benefit of additional data can be substantially offset by the cost due to the changed data ownership distribution. The symptom of this is that the top left to bottom right diagonal, where the forces work against each other, varies much less than the bottom left to top right diagonal, where they reinforce each other. The strength of this effect, and in particular its potential to overturn the benefits of substantial dataset increases, suggests questions beyond this paper's scope. Nevertheless, they are significant as they call for a re-think of our view of data imbalance as a mere convergence speed issue. The leveraging of this observation and its use in the design of superior aggregation strategies is left as future work.

\section{Conclusion}

This study delivered a privacy-preserving distributed learning implementation that both meets the limiting constraints of the industry's regulatory and commercial constraints and outperforms previously available alternatives by up to 15\%. Furthermore, due to its unique data structure, our investigation demonstrated FL in DTI as the first identified data distribution-agnostic domain. Finally, we identified a material trade-off between the benefits of adding new data and the cost of introducing more clients. This observation is of particular relevance as it breaks the generally accepted maxim that more data is always better and thus motivates the need for further exploration to design superior federated learning algorithms. 


\begin{acks}
This work was supported by the UK’s EPSRC with the OPERA (EP/R018677/1) and the MOA (EP/S001530/1) projects; the ERC via the REDIAL project (805194); the Project HPC-EUROPA3 (INFRAIA-2016-1-730897), with the support of the EC Research Innovation Action under the H2020 Programme;  the European PILOT project via the EuroHPC JU (101034126); the CINECA award under the ISCRA initiative; the computer resources and technical support provided by ICHEC; the Spoke "FutureHPC \& BigData” of the ICSC – Centro Nazionale di Ricerca in "High Performance Computing, Big Data and Quantum Computing", funded by European Union – NextGenerationEU.
\end{acks}

\bibliographystyle{ACM-Reference-Format}
\bibliography{bibfile}


\begin{thebibliography}{20}


\ifx \showCODEN    \undefined \def \showCODEN     #1{\unskip}     \fi
\ifx \showDOI      \undefined \def \showDOI       #1{#1}\fi
\ifx \showISBNx    \undefined \def \showISBNx     #1{\unskip}     \fi
\ifx \showISBNxiii \undefined \def \showISBNxiii  #1{\unskip}     \fi
\ifx \showISSN     \undefined \def \showISSN      #1{\unskip}     \fi
\ifx \showLCCN     \undefined \def \showLCCN      #1{\unskip}     \fi
\ifx \shownote     \undefined \def \shownote      #1{#1}          \fi
\ifx \showarticletitle \undefined \def \showarticletitle #1{#1}   \fi
\ifx \showURL      \undefined \def \showURL       {\relax}        \fi
\providecommand\bibfield[2]{#2}
\providecommand\bibinfo[2]{#2}
\providecommand\natexlab[1]{#1}
\providecommand\showeprint[2][]{arXiv:#2}

\bibitem[Beutel et~al\mbox{.}(2020)]%
        {beutel2020flower}
\bibfield{author}{\bibinfo{person}{Daniel~J Beutel}, \bibinfo{person}{Taner
  Topal}, \bibinfo{person}{Akhil Mathur}, \bibinfo{person}{Xinchi Qiu},
  \bibinfo{person}{Titouan Parcollet}, \bibinfo{person}{Pedro~PB de
  Gusm{\~a}o}, {and} \bibinfo{person}{Nicholas~D Lane}.}
  \bibinfo{year}{2020}\natexlab{}.
\newblock \showarticletitle{Flower: A friendly federated learning research
  framework}.
\newblock \bibinfo{journal}{\emph{arXiv preprint arXiv:2007.14390}}
  (\bibinfo{year}{2020}).
\newblock


\bibitem[Breiman(1996)]%
        {breiman1996bagging}
\bibfield{author}{\bibinfo{person}{Leo Breiman}.}
  \bibinfo{year}{1996}\natexlab{}.
\newblock \showarticletitle{Bagging predictors}.
\newblock \bibinfo{journal}{\emph{Machine learning}} \bibinfo{volume}{24},
  \bibinfo{number}{2} (\bibinfo{year}{1996}), \bibinfo{pages}{123--140}.
\newblock


\bibitem[Chen et~al\mbox{.}(2021)]%
        {chen2021fl}
\bibfield{author}{\bibinfo{person}{Shaoqi Chen}, \bibinfo{person}{Dongyu Xue},
  \bibinfo{person}{Guohui Chuai}, \bibinfo{person}{Qiang Yang}, {and}
  \bibinfo{person}{Qi Liu}.} \bibinfo{year}{2021}\natexlab{}.
\newblock \showarticletitle{FL-QSAR: a federated learning-based QSAR prototype
  for collaborative drug discovery}.
\newblock \bibinfo{journal}{\emph{Bioinformatics}} \bibinfo{volume}{36},
  \bibinfo{number}{22-23} (\bibinfo{year}{2021}), \bibinfo{pages}{5492--5498}.
\newblock


\bibitem[Elinas et~al\mbox{.}(2020)]%
        {elinas2020variational}
\bibfield{author}{\bibinfo{person}{Pantelis Elinas}, \bibinfo{person}{Edwin~V
  Bonilla}, {and} \bibinfo{person}{Louis Tiao}.}
  \bibinfo{year}{2020}\natexlab{}.
\newblock \showarticletitle{Variational inference for graph convolutional
  networks in the absence of graph data and adversarial settings}.
\newblock \bibinfo{journal}{\emph{Advances in Neural Information Processing
  Systems}}  \bibinfo{volume}{33} (\bibinfo{year}{2020}),
  \bibinfo{pages}{18648--18660}.
\newblock


\bibitem[Hie et~al\mbox{.}(2018)]%
        {hie2018realizing}
\bibfield{author}{\bibinfo{person}{Brian Hie}, \bibinfo{person}{Hyunghoon Cho},
  {and} \bibinfo{person}{Bonnie Berger}.} \bibinfo{year}{2018}\natexlab{}.
\newblock \showarticletitle{Realizing private and practical pharmacological
  collaboration}.
\newblock \bibinfo{journal}{\emph{Science}} \bibinfo{volume}{362},
  \bibinfo{number}{6412} (\bibinfo{year}{2018}), \bibinfo{pages}{347--350}.
\newblock


\bibitem[Huang et~al\mbox{.}(2020)]%
        {huang2020deeppurpose}
\bibfield{author}{\bibinfo{person}{Kexin Huang}, \bibinfo{person}{Tianfan Fu},
  \bibinfo{person}{Lucas~M Glass}, \bibinfo{person}{Marinka Zitnik},
  \bibinfo{person}{Cao Xiao}, {and} \bibinfo{person}{Jimeng Sun}.}
  \bibinfo{year}{2020}\natexlab{}.
\newblock \showarticletitle{DeepPurpose: a deep learning library for
  drug--target interaction prediction}.
\newblock \bibinfo{journal}{\emph{Bioinformatics}} \bibinfo{volume}{36},
  \bibinfo{number}{22-23} (\bibinfo{year}{2020}), \bibinfo{pages}{5545--5547}.
\newblock


\bibitem[Joshi et~al\mbox{.}(2022)]%
        {joshi2022federated}
\bibfield{author}{\bibinfo{person}{Madhura Joshi}, \bibinfo{person}{Ankit Pal},
  {and} \bibinfo{person}{Malaikannan Sankarasubbu}.}
  \bibinfo{year}{2022}\natexlab{}.
\newblock \showarticletitle{Federated Learning for Healthcare Domain-Pipeline,
  Applications and Challenges}.
\newblock \bibinfo{journal}{\emph{ACM Transactions on Computing for
  Healthcare}} (\bibinfo{year}{2022}).
\newblock


\bibitem[Kairouz et~al\mbox{.}(2021)]%
        {kairouz2021advances}
\bibfield{author}{\bibinfo{person}{Peter Kairouz}, \bibinfo{person}{H~Brendan
  McMahan}, \bibinfo{person}{Brendan Avent}, \bibinfo{person}{Aur{\'e}lien
  Bellet}, \bibinfo{person}{Mehdi Bennis}, \bibinfo{person}{Arjun~Nitin
  Bhagoji}, \bibinfo{person}{Kallista Bonawitz}, \bibinfo{person}{Zachary
  Charles}, \bibinfo{person}{Graham Cormode}, \bibinfo{person}{Rachel
  Cummings}, {et~al\mbox{.}}} \bibinfo{year}{2021}\natexlab{}.
\newblock \showarticletitle{Advances and open problems in federated learning}.
\newblock \bibinfo{journal}{\emph{Foundations and Trends{\textregistered} in
  Machine Learning}} \bibinfo{volume}{14}, \bibinfo{number}{1--2}
  (\bibinfo{year}{2021}), \bibinfo{pages}{1--210}.
\newblock


\bibitem[Li et~al\mbox{.}(2020)]%
        {li2020federated}
\bibfield{author}{\bibinfo{person}{Tian Li}, \bibinfo{person}{Anit~Kumar Sahu},
  \bibinfo{person}{Ameet Talwalkar}, {and} \bibinfo{person}{Virginia Smith}.}
  \bibinfo{year}{2020}\natexlab{}.
\newblock \showarticletitle{Federated learning: Challenges, methods, and future
  directions}.
\newblock \bibinfo{journal}{\emph{IEEE Signal Processing Magazine}}
  \bibinfo{volume}{37}, \bibinfo{number}{3} (\bibinfo{year}{2020}),
  \bibinfo{pages}{50--60}.
\newblock


\bibitem[Ma et~al\mbox{.}(2020)]%
        {ma2020secure}
\bibfield{author}{\bibinfo{person}{Rong Ma}, \bibinfo{person}{Yi Li},
  \bibinfo{person}{Chenxing Li}, \bibinfo{person}{Fangping Wan},
  \bibinfo{person}{Hailin Hu}, \bibinfo{person}{Wei Xu}, {and}
  \bibinfo{person}{Jianyang Zeng}.} \bibinfo{year}{2020}\natexlab{}.
\newblock \showarticletitle{Secure multiparty computation for
  privacy-preserving drug discovery}.
\newblock \bibinfo{journal}{\emph{Bioinformatics}} \bibinfo{volume}{36},
  \bibinfo{number}{9} (\bibinfo{year}{2020}), \bibinfo{pages}{2872--2880}.
\newblock


\bibitem[McMahan et~al\mbox{.}(2017)]%
        {fedAvg}
\bibfield{author}{\bibinfo{person}{Brendan McMahan}, \bibinfo{person}{Eider
  Moore}, \bibinfo{person}{Daniel Ramage}, \bibinfo{person}{Seth Hampson},
  {and} \bibinfo{person}{Blaise Ag{\"{u}}era~y Arcas}.}
  \bibinfo{year}{2017}\natexlab{}.
\newblock \showarticletitle{Communication-Efficient Learning of Deep Networks
  from Decentralized Data}. In \bibinfo{booktitle}{\emph{Proceedings of the
  20th International Conference on Artificial Intelligence and Statistics
  {AISTATS}}} \emph{(\bibinfo{series}{Proceedings of Machine Learning
  Research}, Vol.~\bibinfo{volume}{54})},
  \bibfield{editor}{\bibinfo{person}{Aarti Singh} {and}
  \bibinfo{person}{Xiaojin~(Jerry) Zhu}} (Eds.). \bibinfo{publisher}{PMLR},
  \bibinfo{address}{Fort Lauderdale, FL, USA}, \bibinfo{pages}{1273--1282}.
\newblock
\showISBNx{9781713807933}


\bibitem[Nguyen et~al\mbox{.}(2022)]%
        {nguyen2022federated}
\bibfield{author}{\bibinfo{person}{Dinh~C Nguyen}, \bibinfo{person}{Quoc-Viet
  Pham}, \bibinfo{person}{Pubudu~N Pathirana}, \bibinfo{person}{Ming Ding},
  \bibinfo{person}{Aruna Seneviratne}, \bibinfo{person}{Zihuai Lin},
  \bibinfo{person}{Octavia Dobre}, {and} \bibinfo{person}{Won-Joo Hwang}.}
  \bibinfo{year}{2022}\natexlab{}.
\newblock \showarticletitle{Federated learning for smart healthcare: A survey}.
\newblock \bibinfo{journal}{\emph{ACM Computing Surveys (CSUR)}}
  \bibinfo{volume}{55}, \bibinfo{number}{3} (\bibinfo{year}{2022}),
  \bibinfo{pages}{1--37}.
\newblock


\bibitem[Nguyen et~al\mbox{.}(2021a)]%
        {nguyen2021graphdta}
\bibfield{author}{\bibinfo{person}{Thin Nguyen}, \bibinfo{person}{Hang Le},
  \bibinfo{person}{Thomas~P Quinn}, \bibinfo{person}{Tri Nguyen},
  \bibinfo{person}{Thuc~Duy Le}, {and} \bibinfo{person}{Svetha Venkatesh}.}
  \bibinfo{year}{2021}\natexlab{a}.
\newblock \showarticletitle{GraphDTA: Predicting drug--target binding affinity
  with graph neural networks}.
\newblock \bibinfo{journal}{\emph{Bioinformatics}} \bibinfo{volume}{37},
  \bibinfo{number}{8} (\bibinfo{year}{2021}), \bibinfo{pages}{1140--1147}.
\newblock


\bibitem[Nguyen et~al\mbox{.}(2021b)]%
        {nguyen2021graph}
\bibfield{author}{\bibinfo{person}{Tuan Nguyen}, \bibinfo{person}{Giang~TT
  Nguyen}, \bibinfo{person}{Thin Nguyen}, {and} \bibinfo{person}{Duc-Hau Le}.}
  \bibinfo{year}{2021}\natexlab{b}.
\newblock \showarticletitle{Graph convolutional networks for drug response
  prediction}.
\newblock \bibinfo{journal}{\emph{IEEE/ACM transactions on computational
  biology and bioinformatics}} \bibinfo{volume}{19}, \bibinfo{number}{1}
  (\bibinfo{year}{2021}), \bibinfo{pages}{146--154}.
\newblock


\bibitem[Pfitzner et~al\mbox{.}(2021)]%
        {pfitzner2021federated}
\bibfield{author}{\bibinfo{person}{Bjarne Pfitzner}, \bibinfo{person}{Nico
  Steckhan}, {and} \bibinfo{person}{Bert Arnrich}.}
  \bibinfo{year}{2021}\natexlab{}.
\newblock \showarticletitle{Federated learning in a medical context: a
  systematic literature review}.
\newblock \bibinfo{journal}{\emph{ACM Transactions on Internet Technology
  (TOIT)}} \bibinfo{volume}{21}, \bibinfo{number}{2} (\bibinfo{year}{2021}),
  \bibinfo{pages}{1--31}.
\newblock


\bibitem[Polato et~al\mbox{.}(2022)]%
        {polato2022boosting}
\bibfield{author}{\bibinfo{person}{Mirko Polato}, \bibinfo{person}{Roberto
  Esposito}, {and} \bibinfo{person}{Marco Aldinucci}.}
  \bibinfo{year}{2022}\natexlab{}.
\newblock \showarticletitle{Boosting the Federation: Cross-Silo Federated
  Learning without Gradient Descent}.
\newblock \bibinfo{journal}{\emph{2022 International Joint Conference on Neural
  Networks (IJCNN)}} (\bibinfo{date}{18-23 July} \bibinfo{year}{2022}),
  \bibinfo{pages}{1--10}.
\newblock
\showISSN{2161-4407}
\urldef\tempurl%
\url{https://doi.org/10.1109/IJCNN55064.2022.9892284.}
\showDOI{\tempurl}


\bibitem[Tang et~al\mbox{.}(2014)]%
        {kiba}
\bibfield{author}{\bibinfo{person}{Jing Tang}, \bibinfo{person}{Agnieszka
  Szwajda}, \bibinfo{person}{Sushil Shakyawar}, \bibinfo{person}{Tao Xu},
  \bibinfo{person}{Petteri Hintsanen}, \bibinfo{person}{Krister Wennerberg},
  {and} \bibinfo{person}{Tero Aittokallio}.} \bibinfo{year}{2014}\natexlab{}.
\newblock \showarticletitle{Making sense of large-scale kinase inhibitor
  bioactivity data sets: a comparative and integrative analysis}.
\newblock \bibinfo{journal}{\emph{Journal of Chemical Information and
  Modeling}} \bibinfo{volume}{54}, \bibinfo{number}{3} (\bibinfo{year}{2014}),
  \bibinfo{pages}{735--743}.
\newblock


\bibitem[Topol(2019)]%
        {topol2019high}
\bibfield{author}{\bibinfo{person}{Eric~J Topol}.}
  \bibinfo{year}{2019}\natexlab{}.
\newblock \showarticletitle{High-performance medicine: the convergence of human
  and artificial intelligence}.
\newblock \bibinfo{journal}{\emph{Nature medicine}} \bibinfo{volume}{25},
  \bibinfo{number}{1} (\bibinfo{year}{2019}), \bibinfo{pages}{44--56}.
\newblock


\bibitem[Wei et~al\mbox{.}(2020)]%
        {wei2020federated}
\bibfield{author}{\bibinfo{person}{Kang Wei}, \bibinfo{person}{Jun Li},
  \bibinfo{person}{Ming Ding}, \bibinfo{person}{Chuan Ma},
  \bibinfo{person}{Howard~H Yang}, \bibinfo{person}{Farhad Farokhi},
  \bibinfo{person}{Shi Jin}, \bibinfo{person}{Tony~QS Quek}, {and}
  \bibinfo{person}{H~Vincent Poor}.} \bibinfo{year}{2020}\natexlab{}.
\newblock \showarticletitle{Federated learning with differential privacy:
  Algorithms and performance analysis}.
\newblock \bibinfo{journal}{\emph{IEEE Transactions on Information Forensics
  and Security}}  \bibinfo{volume}{15} (\bibinfo{year}{2020}),
  \bibinfo{pages}{3454--3469}.
\newblock


\bibitem[Wu and Wang(2021)]%
        {wu2021fast}
\bibfield{author}{\bibinfo{person}{Hongda Wu} {and} \bibinfo{person}{Ping
  Wang}.} \bibinfo{year}{2021}\natexlab{}.
\newblock \showarticletitle{Fast-convergent federated learning with adaptive
  weighting}.
\newblock \bibinfo{journal}{\emph{IEEE Transactions on Cognitive Communications
  and Networking}} \bibinfo{volume}{7}, \bibinfo{number}{4}
  (\bibinfo{year}{2021}), \bibinfo{pages}{1078--1088}.
\newblock


\end{thebibliography}

\end{document}